\documentclass[runningheads]{llncs}
\usepackage{graphicx}
\usepackage{multirow}
\usepackage{subfigure}
\usepackage{url}
\usepackage{epstopdf}
\usepackage{color}
\usepackage{xcolor}
\usepackage{bbm}
\usepackage[cmex10]{amsmath}
\usepackage{amssymb}
\usepackage{array}
\usepackage{dsfont}
\usepackage{caption}
\usepackage{multirow}
\usepackage{url}
\usepackage{bbding}

\begin{document}
\title{Increased-confidence Adversarial Examples for Deep Learning Counter-Forensics}
\titlerunning{Increased-confidence Adversarial Examples for DL Counter-Forensics}
\author{Wenjie Li\inst{1,2} \and
Benedetta Tondi\inst{3} \and
Rongrong Ni\inst{1,2} \Envelope \and
Mauro Barni\inst{3}
}
\authorrunning{Li \emph{et al.}}
\institute{Institute of Information Science, Beijing Jiaotong University, Beijing, 100044, China \and
Beijing Key Laboratory of Advanced Information Science and Network Technology, Beijing, 100044, China \\ \email{wenjie\_li@bjtu.edu.cn, rrni@bjtu.edu.cn}\\ \and
Department of Information Engineering and Mathematical Sciences,\\ University of Siena, Italy \\
\email{benedettatondi@gmail.com, barni@dii.unisi.it}}
\maketitle              % typeset the header of the contribution
\begin{abstract}
Transferability of adversarial examples is a key issue to apply this kind of attacks against multimedia forensics (MMF) techniques based on Deep Learning (DL) in a real-life setting. Adversarial example transferability, in fact, would open the way to the deployment of successful counter forensics attacks also in cases where the attacker does not have a full knowledge of the to-be-attacked system. Some preliminary works have shown that adversarial examples against CNN-based image forensics detectors are in general non-transferrable, at least when the basic versions of the attacks implemented in the most popular libraries are adopted. In this paper, we introduce a general strategy to increase the strength of the attacks and evaluate their transferability when such a strength varies. We experimentally show that, in this way, attack transferability can be largely increased, at the expense of a larger distortion. Our research confirms the security threats posed by the existence of adversarial examples even in multimedia forensics scenarios, thus calling for new defense strategies to improve the security of DL-based MMF techniques.

\end{abstract}
\section{Introduction}\label{sec:intro}
Several studies in machine learning have shown that adversarial attacks carried out against a given network (in most cases, but not only, a convolutional neural network - CNN) are often transferable to other networks designed for the same task, that is, the attacks maintain part of their effectiveness even against networks other than that considered to construct the attack \cite{PaperTransf16,TramSpAdv17}.
Attack transferability is a key property, especially in security-related scenarios like multimedia forensics, since it opens the way towards the deployment of powerful attacks to be used in real-life applications, wherein the attacker does not have full access to the attacked network (gray-box scenario). Thanks to transferability, in fact, the attacker can attack a surrogate network mimicking the target one and the attack will be effective against the target network with a large probability \cite{barni2018adversarial,PaperTransf16}.

While attack transferability was first proved for computer vision applications, recent research has shown that common adversarial examples are often non-transferable in image forensic applications \cite{barni2019transferability,Gragnaniello2018AnalysisOA,VerdoAdv18}. Results like those reported in \cite{barni2019transferability,Gragnaniello2018AnalysisOA,VerdoAdv18} represent a great challenge for attackers who want to exploit the existence of adversarial examples to attack MMF systems in a real life setting wherein the controlled conditions typical of laboratory experiments  do not hold. The problem is exacerbated by the fact that the most popular attack software packages \cite{papernot2018cleverhans,rauber2017foolbox} are designed in such a way to minimize the embedding distortion for a successful attack. In this way, the attacked samples are often very close to the decision boundary, so that even a slight modification to the detector may undermine the effectiveness of the attack. On the contrary, the attacker could prefer to introduce a larger distortion, if doing so permits him to obtain an attacked sample that lies deeper into the target region of the attack (since in this way the probability that the attack can be transferred to another network increases). Unfortunately, controlling the distance of the attacked samples to the decision boundary, and hence controlling their resilience to perturbations of the boundary is not easy, given the complexity of the decision boundary learned by CNNs.

In this paper, we introduce a general strategy to increase the strength of the attacks, and evaluate the transferability of the adversarial examples generated in this way in a multimedia forensics context.
Specifically, we show that stronger attacks can be built by increasing the {\em confidence} of the misclassification, namely the difference between the score (logits) of the target class of the attack and the highest score of the other classes.
Our experiments reveal that controlling the strength of the attack by varying its confidence level is an effective way to improve the transferability of the attack, at the expense of a slightly larger distortion.

The rest of the paper is organized as follows. In Sect. \ref{sec.attackForm}, the proposed confidence-controlled attacks generating adversarial examples with improved transferability is described. In Sect. \ref{sec.method}, the methodology used for the evaluation of transferability is introduced. Then, the experimental results as well as corresponding discussions are given in Sect. \ref{sec.exp}. Finally, Sect. \ref{sec:conclusion} concludes the paper.

\section{Proposed Confidence-Controlled Attacks}
\label{sec.attackForm}

The most straightforward way of controlling the strength of the attacks (at least those based on gradients) would be to go on with the attack iterations until a limit value of the PSNR (Peak Signal to Noise Ratio) is reached. The PSNR, then, would give a measure of the attack strength. Some preliminary experiments we have carried out, however, reveal that controlling the strength of the attack in this way is not easy. One reason for such a difficulty is the intricacy of the decision boundary, so that a lower PSNR does not necessarily result in a stronger attack. In addition, if the gradient is computed directly with respect to the final soft output of the network, as soon as we depart from the decision boundary the gradient tends to vanish, thus making it difficult for the attacker to find a good descent direction ensuring a larger distance from the decision boundary.

To cope with the above difficulties, we have devised a solution that modifies the stop condition of the attack in such a way to control the confidence of the misclassification induced by the attack. In this way, the solver of the optimization problem the attack relies on is forced to find an adversarial example that is classified as the target class with  the {\em desired}  minimum confidence. Otherwise,  a failure is declared (no adversarial examples can be returned with the desired confidence).

In the following, we describe the confidence-based attack focusing on the case of binary classification\footnote{While trivial, the extension to multi-class classification requires to distinguish between targeted and non-targeted attacks.}.
Let $\phi$ be the function describing the output of the neural network and let $y$ denote the class label, $y=0,1$.
Let $z_i$  be the logits, i.e. the outputs of the neural network prior to the softmax layer, corresponding to the two labels $i=0,1$. The final value $\phi$ is obtained by applying the softmax to the logits and thresholding the output. Given an image $X$ such that $y=i$ ($i=0,1$), an image $X'$ is declared adversarial {\em if and only if} $z_{1-i} - z_i > c$, where $c > 0$ is the desired {\em confidence}.
All the most popular adversarial attack algorithms (e.g. FGSM, PGD) can be modified by generalizing the stop condition in this way.
%considering the above generalization.
The goal of our research is to evaluate the transferability of adversarial attacks against multimedia forensics detectors, as a function of $c$.

Several other methods have been proposed to improve the transferability of adversarial examples.
The regularization-based method proposed in \cite{TAP}, for instance, improves transferability by adding a regularization term to the loss function to smooth the perturbation.
Ensemble-based methods aim at finding an universal perturbation which is effective for an ensemble of architectures \cite{tramr2017ensemble,liu2016delving} or input images \cite{Dong-2019-CVPR,DI-IFGSM}. In particular, the method recently proposed in \cite{DI-IFGSM}, called Diverse Inputs I-FGSM ($\text{DI}^2\text{-FGSM}$) increases the transferability of the attacks by exploiting input diversity.
This method is an extension of the basic I-FGSM, and it has been proven to significantly improve the transferability of adversarial examples by applying random transformations (random resizing and random padding) to the input images at each iteration. Given its effectiveness, in Sect. \ref{sec.exp}, the transferability of increased-confidence adversarial examples is compared with that achieved by $\text{DI}^2\text{-FGSM}$.

\section{Methodology}
\label{sec.method}

In order to carry out a comprehensive investigation, we assessed the attack transferability under various sources and degrees of mismatch between the {\em source network}, i.e., the network used to perform the adversarial attack (SN), and the {\em target  network}, i.e., the one the attack
should be transferred to (TN).
In particular, we considered the following types of transferability: i) cross-network (different network architectures trained on the same dataset),
ii) cross-training (the same architecture trained on different datasets) and iii) cross-network-and-training (different architectures trained on different datasets).
We carried out our experiments with different
kinds of attacks, considering three manipulation detection tasks (median filtering, image resizing and additive white Gaussian noise (AWGN)) and three
different architectures, as described below.

\subsection{Attacks}

We created the adversarial examples by applying the confidence-based stop condition to the following methods:
i) Iterative Fast Gradient Sign Method (I-FGSM) \cite{goodfellow2014explaining}, i.e. the refined iterative version of the original FGSM attack,
ii) the Projected Gradient Descent (PGD) attack \cite{kurakin2016adversarial,madry2017towards},
iii) C\&W attack \cite{CW2017},
iv) Momentum Iterative FGSM  (MI-FGSM) \cite{Dong_MI-FGSM}.

A brief description of the basic version of the above attacks is provided in the following. With I-FGSM, at each iteration, an adversarial
perturbation is obtained by computing the gradient
of the loss function with respect to the input image and considering its
sign multiplied by a (normalized) strength factor $\varepsilon$.
The algorithm is applied iteratively until an adversarial image is obtained (that is, an image which is misclassified by the network), for a maximum number of steps.
PGD looks for the perturbation that maximizes the loss function under some restrictions regarding the introduced $L_{\infty}$ distortion.
Specifically, at each iteration,  the image is first updated (similarly to I-FGSM)
by using a certain value of $\varepsilon$;
then, the pixel values are clipped to ensure that they remain in the $\alpha$-neighbourhood of the original image.
The C\&W attack based on the $L_2$ metric
minimizes  a loss function that weighs a classification loss
with a distance term measuring the distortion between the original and the attacked images, after applying a $\tanh$-nonlinearity to enforce the attacked samples to take valid values in a bounding box.
MI-FGSM uses the momentum method to stabilize the update directions.
It works by adding the accumulated gradients from previous steps to the current gradient with a decay factor $\mu$.
In this way, the optimization should escape from poor local maxima/minima during the iterations
(in principle, the momentum iterative method can be easily applied on top of any iterative attack algorithms. By following \cite{Dong_MI-FGSM}, we considered I-FGSM).

Some observations are in order.
C\&W method already introduces a confidence parameter, named $\kappa$, in its formulation of the objective function,
so that the solver is `encouraged' to find an adversarial example that belongs to the desired class with high confidence (smaller value of the objective function).
However, by running the attack with a standard implementation, it turns out that the confidence achieved by the attacked image is not always $\kappa$ (depending on the number of iterations considered for the attack and other parameters \cite{CW2017}). Then, we applied our confidence-based stop condition  to this attack as well, to ensure that the desired confidence is always achieved (with $c = \kappa$).
Another observation regards MI-FGSM. Integrating the momentum term into the iterative process is something that should by itself increase the transferability of the adversarial examples \cite{Dong_MI-FGSM}.
Then, in this case, by considering $c> 0$, we are basically combining together two ways to get a stronger attack.
However, according to our results, the improvement of transferability that can be obtained by using the momentum is a minor one, thus justifying the further adoption of our confidence-based strategy.

\vspace{-0.2cm}

\subsection{Datasets and Networks}

For the experiments with cross-training mismatch, we considered the RAISE (R) \cite{RAISE8K} and the VISION (V) datasets \cite{Shull17}.
In particular, a number of 2000 uncompressed, camera-native images (.tiff) with size of $4288\times2848$ were taken from the RAISE dataset.
The same number of images were taken from the VISION dataset.
This dataset consists of native images acquired by smartphones/tablets belonging to several different brands.
%To get images with similar resolution, we selected only the devices for which the resolution was not very different from that of RAISE images.
To get images with similar resolution, we selected only the devices with the image resolution close to $4288\times2848$.
The native images from both R and V datasets were split into training, validation and test sets, and then processed to get the manipulated images, namely, median-filtered (by a 5 $\times$ 5 window), resized (downsampling, by a factor of 0.8) and noisy (Gaussian noise with std dev 1) images.
We considered gray-level images for our experiments, then all the RGB images were first converted to gray-scale.

Concerning the network architectures used to build the detectors, we considered the network in \cite{bayar2016deep} from Bayar and Stamm (BSnet),
and the one in \cite{BCNT18} from Barni and Costanzo \emph{et al.} (BC+net).
BSnet was originally designed for standard manipulation detection tasks and it is not very deep (3 convolutional layers, 3 max-pooling layers, and 2 fully connected layers).
As the main feature of BSnet, the filters used in the first layer (with $5 \times 5$ receptive field) are forced to have a high-pass nature, so to extract the residual-based features from images. For more details on the network architecture, we refer to \cite{bayar2016deep}.
As for the BC+net, it was initially proposed for the more difficult task of generic contrast adjustment detection.
This network is much deeper (9 convolutional layers) than BSnet, and no constraint is applied to the filters used in the network.
More details can be found in the original paper \cite{BCNT18}.
In addition to these two networks designed for image forensics tasks, we also considered a common network used for pattern recognition applications, that is, the VGG-16 network (VGGnet) \cite{VGG}.
The architecture considered consists of 13 convolutional layers in total, and, respectively, 1024, 256, and 2 (output) neurons for the three fully connected layers (see \cite{VGG} for further information).
The clear notations for the models used in the experiments are given below. For a given detection task,
we indicate the trained model as $\phi_{\text{Net}}^{\text{DB}}$, where ``Net" indicates the architecture with ``Net"  $\in \{\text{BSnet, BC+net, VGGnet}\}$, and ``DB" $\in \{\text{R, V}\}$ is the dataset used for training.

\vspace{-0.2cm}

\section{Experiments}
\label{sec.exp}

\subsection{Setup}

Based on the three different CNN architectures described in the previous section, we trained the models in the following way.
To build the BSnet models for the three detection tasks, we considered $2 \times 10^5$ patches per class for training (and validation) and $10^4$ for testing.
For the VGGnet models, a number of $10^5$ patches per class was used for training and validation, and the test set consists of $10^4$ patches.
A maximum number of 100 patches was (randomly) selected from each image to increase patch diversity by using more images. Finally, for the BC+net's,  we considered $10^6$ patches for training, $10^5$ patches for validation and $5 \times 10^4$ patches for testing.
To reach these numbers, we selected all the non-overlapping patches from each image.
For all the models, the patch size was set to $128 \times 128$. The batch size for training was set to 32 patches.
For BSnet and BC+net, the Adam solver with learning rate $10^{-4}$ was used. The learning rate was set to $10^{-5}$ for VGGnet. The number of epochs was set, respectively, to 30 for BSnet \cite{bayar2016deep}, 50 for VGGnet,  and 3 for BC+net (following \cite{BCNT18}).
The accuracies achieved by these models on the test sets range between 98.1\% and 99.5\%  for median filtering detection, from 96.6\% to 99.0\% for resizing detection, and from 98.3\% to 99.9\% for the detection of AWGN.

We attacked images from the manipulated class only. In counter-forensic applications, in fact, it is reasonable to assume that the attack is conducted only in one direction, since the attacker wants to pass off a manipulated image as an original one to cause a false negative error.
Performance are measured on 500 attacked images, obtained by attacking a subset of manipulated images from the test set.
The Foolbox toolbox \cite{rauber2017foolbox} was used to implement the attacks.
The parameters of the attacks were set as follows\footnote{For simplicity, we report only the values of the Foolbox input parameters for each attack.
For their meaning and the correspondences with the parameters of the attacks, we refer to \cite{rauber2017foolbox}.}.
For C\&W, all the parameters were set to the default values.
For PGD, we set `epsilon' = 0.3, `stepsize' = 0.005 and `iterations' = 100.
A binary search is performed
over `epsilon' and `stepsize' to optimize the choice of the hyperparameters,
using the input values only for the initialization of the algorithm.
For I-FGSM, we set `epsilons' = 10 and `steps' = 100.
For MI-FGSM, the parameters used in I-FGSM were also adopted and the decay factor of the accumulated gradient was set to $\mu = 0.2$ (similar results were obtained with lower values of $\mu$).
For the comparison with the $\text{DI}^2\text{-FGSM}$, we followed the parameter setting in \cite{DI-IFGSM},
where $\varepsilon = 1/255$. We decreased this value to $0.002$ in the case of AWGN detection, for which letting $\varepsilon = 1/255$ results in a very low attack success rate.
Besides, the input diversity is achieved by resizing the image to $r\times r$ with $r\in[100,128)$ and then randomly padding it to the size of $128\times128$, and the transformation probability is also set to 0.5, which is the same with that used in \cite{DI-IFGSM}.

Several mismatch combinations of SN and TN were considered for all detection tasks, and the transferability of the adversarial examples for the cases of cross-network, cross-training and cross-network-and-training is investigated. In the following, we will only report the results for a subset of these cases, which is sufficient to draw the most significant conclusions.

\subsection{Results}
For each attack method applied with the proposed confidence-based stop condition, we report the attack success rate with respect to the target network, ASR$_{\text{TN}}$, and the PSNR of the attacked images averaged on the successfully attacked samples, for several confidence values $c$ ranging from 0 to a maximum value chosen in such a way that the average PSNR is not too low (around 30 dB).
The range of values of $c$ depends on the magnitude of the logits and then on the SN.
The attack success rate with respect to the source network, ASR$_{\text{SN}}$, is $100\%$ or very close to this value for the confidence-based attacks in all cases, then it is not reported.
The performance of the $\text{DI}^2\text{-FGSM}$, including ASR$_{\text{SN}}$, are reported for different numbers of iterations (for a similar PSNR limitation) and discussed at the end of this section.

\begin{table}[t]
	\renewcommand\arraystretch{1.15}
	\centering
	\setlength{\tabcolsep}{6.5pt}
	\caption{Results for cross-network mismatch for the median filtering detection.}
	{\begin{tabular}
			{|m{0.4cm}<{\centering}|m{0.45cm}<{\centering}m{0.45cm}<{\centering}|m{0.45cm}<{\centering}m{0.45cm}<{\centering}|m{0.45cm}<{\centering}m{0.45cm}<{\centering}|m{0.45cm}<{\centering}m{0.45cm}<{\centering}||m{0.35cm}<{\centering}|m{0.44cm}<{\centering}m{0.44cm}<{\centering}m{0.44cm}<{\centering}|}
			\hline
\multicolumn{13}{|c|}{SN = $\phi_{\text{VGGnet}}^{\text{R}}$,  TN = $\phi_{\text{BSnet}}^{\text{R}}$}  \\ \hline
			&\multicolumn{2}{c|}{C\&W} &\multicolumn{2}{c|}{PGD} &\multicolumn{2}{c|}{I-FGSM} &\multicolumn{2}{c||}{MI-FGSM} &\multicolumn{4}{c|}{$\text{DI}^2\text{-FGSM}$}\\ \hline
			$c$  &\tiny{{ASR}$_{\text{TN}}$}    &\tiny{PSNR}   &\tiny{{ASR}$_{\text{TN}}$}    &\tiny{PSNR}   &\tiny{{ASR}$_{\text{TN}}$}    &\tiny{PSNR}     &\tiny{{ASR}$_{\text{TN}}$}    &\tiny{PSNR}  &\scriptsize{iter} &\tiny{{ASR}$_{\text{SN}}$} &\tiny{{ASR}$_{\text{TN}}$}    &\tiny{PSNR}  \\ \hline
			0       &8.4   &69.1    &5.4  &67.5    &27.2  &58.1    &27.2  &58.1  &1  &86.4  &51.8  &48.3 \\ \hline
			12      &50.6  &54.6    &55.0 &52.0    &71.0  &47.9    &71.6  &47.8  &2  &97.0  &69.2  &45.0\\ \hline
		    12.5    &70.1  &51.5    &79.0 &48.9    &88.2  &45.3    &88.6  &45.3  &3  &99.2  &77.6  &43.2\\ \hline
			13      &91.2  &48.1    &94.6 &45.4    &96.4  &42.5    &96.6  &42.7  &5  &100   &86.8  &41.4\\ \hline
\multicolumn{13}{|c|}{SN = $\phi_{\text{BSnet}}^{\text{R}}$,  TN = $\phi_{\text{BC+net}}^{\text{R}}$}  \\ \hline
&\multicolumn{2}{c|}{C\&W} &\multicolumn{2}{c|}{PGD} &\multicolumn{2}{c|}{I-FGSM} &\multicolumn{2}{c||}{MI-FGSM} &\multicolumn{4}{c|}{$\text{DI}^2\text{-FGSM}$}\\ \hline
			$c$  &\tiny{{ASR}$_{\text{TN}}$}    &\tiny{PSNR}   &\tiny{{ASR}$_{\text{TN}}$}    &\tiny{PSNR}   &\tiny{{ASR}$_{\text{TN}}$}    &\tiny{PSNR}     &\tiny{{ASR}$_{\text{TN}}$}    &\tiny{PSNR}  &\scriptsize{iter} &\tiny{{ASR}$_{\text{SN}}$} &\tiny{{ASR}$_{\text{TN}}$}    &\tiny{PSNR}  \\ \hline
			0        &0.2    &72.0    &0.2  &74.5    &23.2  &59.7    &23.4  &59.7  &1  &91.4  &52.4  &48.2\\ \hline
			50       &56.0   &52.2    &55.6 &50.3    &60.6  &48.9    &60.4  &48.9  &2  &98.4  &69.4  &44.6\\ \hline
			80       &74.0   &47.8    &74.0 &45.8    &77.8  &45.1    &78.8  &44.8  &3  &99.4  &77.0  &42.3\\ \hline
			100      &83.6   &45.2    &83.6 &43.5    &85.4  &42.9    &86.2  &42.7  &5  &100   &85.4  &40.2\\ \hline
	\end{tabular}}
	\label{tab:crossmodel1}
\vspace{-0.2cm}
\end{table}
\begin{table}
	\renewcommand\arraystretch{1.15}
	\centering
	\setlength{\tabcolsep}{6.5pt}
	\caption{Results for cross-network mismatch for the image resizing detection.
}
	{\begin{tabular}			{|m{0.4cm}<{\centering}|m{0.45cm}<{\centering}m{0.45cm}<{\centering}|m{0.45cm}<{\centering}m{0.45cm}<{\centering}|m{0.45cm}<{\centering}m{0.45cm}<{\centering}|m{0.45cm}<{\centering}m{0.45cm}<{\centering}||m{0.35cm}<{\centering}|m{0.44cm}<{\centering}m{0.44cm}<{\centering}m{0.44cm}<{\centering}|}
			\hline
\multicolumn{13}{|c|}{SN = $\phi_{\text{VGGnet}}^{\text{R}}$,  TN = $\phi_{\text{BSnet}}^{\text{R}}$}  \\ \hline
			&\multicolumn{2}{c|}{C\&W} &\multicolumn{2}{c|}{PGD} &\multicolumn{2}{c|}{I-FGSM} &\multicolumn{2}{c||}{MI-FGSM} &\multicolumn{4}{c|}{$\text{DI}^2\text{-FGSM}$}\\ \hline
			$c$  &\tiny{{ASR}$_{\text{TN}}$}    &\tiny{PSNR}   &\tiny{{ASR}$_{\text{TN}}$}    &\tiny{PSNR}   &\tiny{{ASR}$_{\text{TN}}$}    &\tiny{PSNR}     &\tiny{{ASR}$_{\text{TN}}$}    &\tiny{PSNR}  &\scriptsize{iter} &\tiny{{ASR}$_{\text{SN}}$} &\tiny{{ASR}$_{\text{TN}}$}    &\tiny{PSNR}  \\ \hline
			0        &1.2    &71.5    &1.8  &75.4      &0.4   &59.3   &0.4   &59.2   &1   &30.6  &2.6   &48.2  \\ \hline
			17       &40.4   &36.8    &22.0 &33.4      &25.0  &33.3   &24.8  &33.3   &25  &100   &8.6   &32.6  \\ \hline
			18       &53.6   &34.3    &39.8 &30.9      &39.6  &30.9   &37.4  &30.9   &35  &100   &23.4  &30.3  \\ \hline
			19       &64.4   &32.2    &52.0 &28.9      &51.6  &28.9   &52.6  &28.9   &45  &100   &41.0  &28.2 \\ \hline
\multicolumn{13}{|c|}{SN = $\phi_{\text{BSnet}}^{\text{R}}$,  TN = $\phi_{\text{BC+net}}^{\text{R}}$}
\\ \hline
&\multicolumn{2}{c|}{C\&W} &\multicolumn{2}{c|}{PGD} &\multicolumn{2}{c|}{I-FGSM} &\multicolumn{2}{c||}{MI-FGSM} &\multicolumn{4}{c|}{$\text{DI}^2\text{-FGSM}$}\\ \hline
			$c$  &\tiny{{ASR}$_{\text{TN}}$}    &\tiny{PSNR}   &\tiny{{ASR}$_{\text{TN}}$}    &\tiny{PSNR}   &\tiny{{ASR}$_{\text{TN}}$}    &\tiny{PSNR}     &\tiny{{ASR}$_{\text{TN}}$}    &\tiny{PSNR}  &\scriptsize{iter} &\tiny{{ASR}$_{\text{SN}}$} &\tiny{{ASR}$_{\text{TN}}$}    &\tiny{PSNR}  \\ \hline
			0        &0.4    &68.3    &0.4  &66.9    &0.4   &58.7    &0.4   &58.6  &1   &96.8  &0.2   &48.2\\ \hline
			50       &82.4   &45.9    &65.2 &42.3    &66.2  &41.9    &63.6  &41.7  &3   &100   &33.8  &41.2\\ \hline
			80       &85.2   &39.3    &84.0 &35.5    &82.4  &35.3    &84.6  &35.4  &8   &100   &77.0  &35.3\\ \hline
			100      &80.4   &34.0    &82.8 &31.6    &83.8  &31.6    &82.2  &31.7  &15  &100   &87.6  &31.1\\ \hline
	\end{tabular}}
	\label{tab:crossmodel2}
\end{table}

\begin{figure*}[t]
  \centering
  \includegraphics[width=0.7\linewidth]{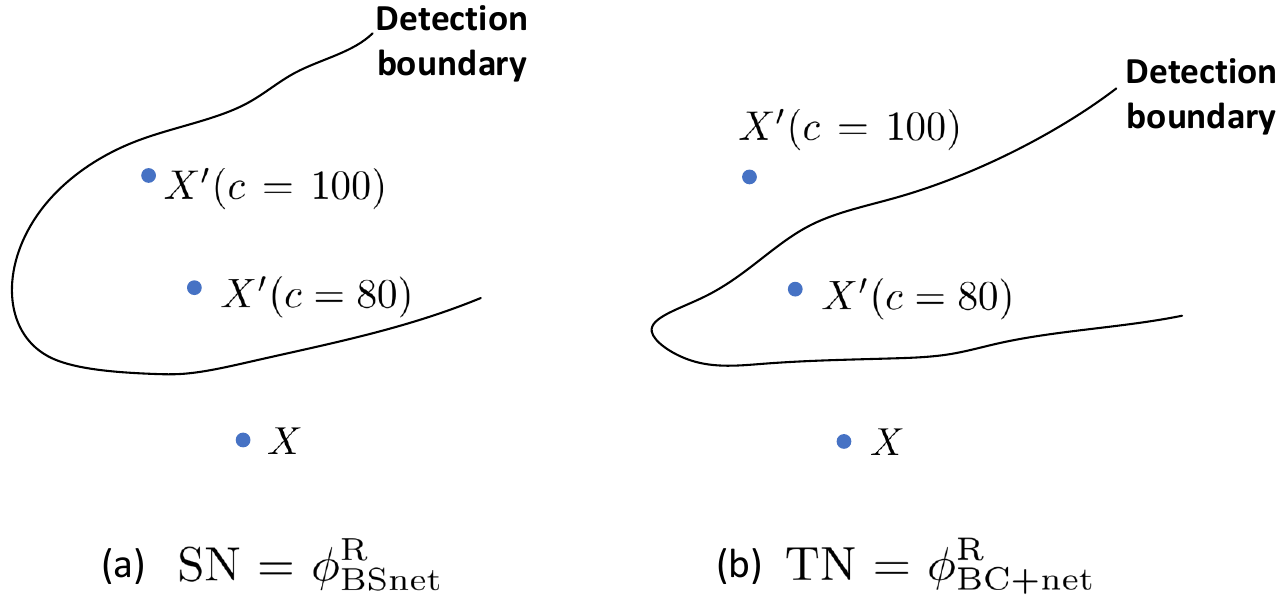}\\
  \caption{An explanation for the phenomenon that a larger $c$ results in less transferability for the case (SN,TN) = $(\phi_{\text{BSnet}}^{\text{R}}, \phi_{\text{BC+net}}^{\text{R}})$ applied to image resizing detection.}
  \label{fig:boundary}
\vspace{-0.2cm}
\end{figure*}

Tables 	\ref{tab:crossmodel1} and \ref{tab:crossmodel2}  show the results for the case of cross-network transferability for two different combinations of SN and TN, i.e., (SN,TN) = $(\phi_{\text{VGGnet}}^{\text{R}}, \phi_{\text{BSnet}}^{\text{R}})$ and (SN,TN) = $(\phi_{\text{BSnet}}^{\text{R}}, \phi_{\text{BC+net}}^{\text{R}})$, for the median filtering and image resizing detection tasks, respectively.
By inspecting the tables, we see that
the transferability of the attacks can be significantly improved in all the cases by increasing the confidence; the ASR$_{\text{TN}}$ passes from
[0-27]\% with $c=0$ to [84-97]\% with the maximum $c$ for median filtering, and from  [0-2]\% with $c=0$ to [52-84]\% with the maximum $c$ for image resizing.
In all cases, the PSNR of the attacked images remains pretty high (larger than 40 dB for the case of median filtering, and about 30 dB for the case of image resizing).
For different SN, the relationship between the values of $c$ and PSNR is different since the exact values assumed by the logits is affected by many factors, such as the network architecture and the detection task.
As a general trend, a larger $c$ always results in adversarial examples with lower PSNR values and higher transferability ({ASR}$_{\text{TN}}$).
The few exceptions to this behaviour, for which the values in Table \ref{tab:crossmodel2} report a lower transferability when using a larger $c$, can be possibly explained by the fact that some perturbed images move too far from the original image and are correctly classified by the TN, although they are still valid adversarial examples for the SN.
A simple illustration of such a behaviour is given in Fig. \ref{fig:boundary} for the case (SN,TN) = $(\phi_{\text{BSnet}}^{\text{R}}, \phi_{\text{BC+net}}^{\text{R}})$ for image resizing detection, where $X' (c=100)$ is adversarial for the SN but is classified correctly by the TN (the distortion is so large that the detection boundary is crossed twice, thus resulting in a correct label).
Moreover, according to the tables, with similar ASR$_{\text{TN}}$ values, the C\&W attack which aims to generate adversarial examples with less distortion can always achieve higher PSNR values than other attack methods.
In summary, increasing the confidence is helpful for the improvement of the transferability of the adversarial examples in these cases.
However, there also exists some combinations of (SN,TN), for which the transferability does not increase much by increasing the confidence value $c$. For instance, when (SN,TN) = $(\phi_{\text{BC+net}}^{\text{R}}, \phi_{\text{BSnet}}^{\text{R}})$ for the image resizing detection task, the ASR$_{\text{TN}}$ can only be increased from [0-2]\% up to around 20\%, under a certain limitation of the PSNR.
This happens when the gradient is almost vanishing,
since in this case a larger confidence does not help in increasing the transferability with respect to the target network.

\begin{figure*}[t]
  \centering
  \includegraphics[width=0.7\linewidth]{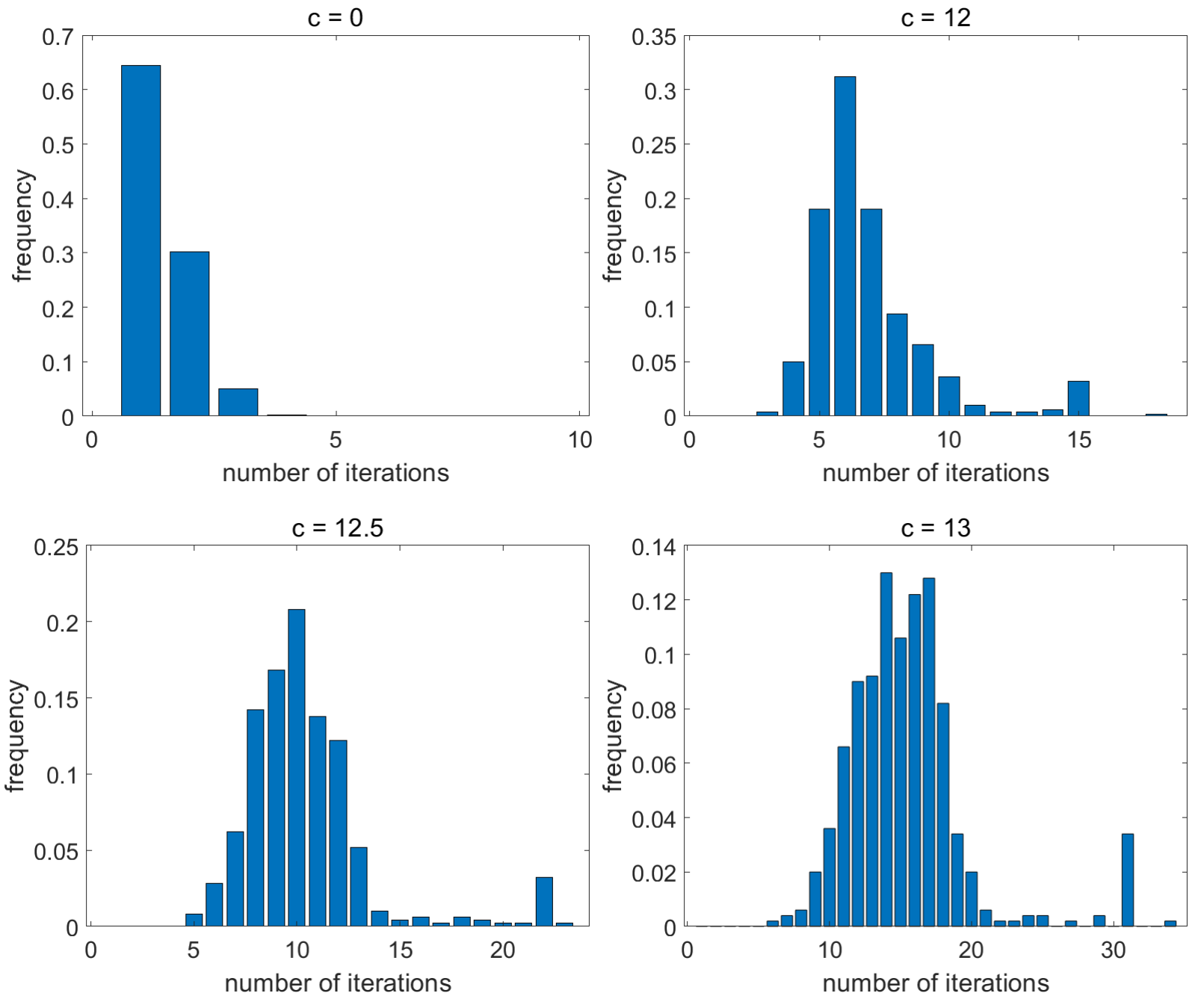}\\
  \caption{Distribution of the number of iterations for the I-FGSM attack using different values of the confidence $c$. The attacked network is $\phi_{\text{VGGnet}}^{\text{R}}$ trained for the median filtering detection task.}
  \label{fig:iterations}
\vspace{-0.2cm}
\end{figure*}

Noticeably, increasing the confidence $c$ permits to increase the transferability also for the MI-FGSM attack, where we did not notice any significant transferability improvement for the case of $c=0$, with respect to the I-FGSM.
A possible explanation is that the strength of the perturbation applied at each iteration of the I-FGSM attack is very small,
then the gradients between subsequent iterations are highly correlated, thus reducing the impact of the gradient stabilization sought for by MI-FGSM. On the other hand, as already pointed out in previous works \cite{barni2019transferability,tondi2018pixel}, if we increase the strength of the perturbation applied at each iteration,
the ASR$_{\text{SN}}$ tends to decrease significantly.
We also observe that the degree of transferability depends on the dataset and the SN architecture, as it was already noticed in \cite{barni2019transferability}.
However, in most cases, the adversarial examples can be successfully transferred by increasing the confidence, and the image quality can be preserved at the same time.
To better investigate the effect of the confidence margin on the attack, we also checked the number of iterations used by the attacks for various confidence values.
Specifically, the I-FGSM attack carried out against $\phi_{\text{VGGnet}}^{\text{R}}$ trained for the median filtering detection task is taken into consideration. The distributions of the number of iterations for different confidence margins are shown in Fig. \ref{fig:iterations}.
According to this figure, when $c=0$, only one or two steps are enough for a successful attack. Then, by increasing $c$, the attack requires more and more iterations. A similar behavior can be observed in the other cases.

\begin{table}[t]
	\renewcommand\arraystretch{1.15}
	\centering
	\setlength{\tabcolsep}{6.5pt}
	\caption{Results for cross-training mismatch for the median filtering detection.}
	{\begin{tabular}
{|m{0.4cm}<{\centering}|m{0.45cm}<{\centering}m{0.45cm}<{\centering}|m{0.45cm}<{\centering}m{0.45cm}<{\centering}|m{0.45cm}<{\centering}m{0.45cm}<{\centering}|m{0.45cm}<{\centering}m{0.45cm}<{\centering}||m{0.35cm}<{\centering}|m{0.44cm}<{\centering}m{0.44cm}<{\centering}m{0.44cm}<{\centering}|}
			\hline
\multicolumn{13}{|c|}{SN = $\phi_{\text{BSnet}}^{\text{R}}$,  TN = $\phi_{\text{BSnet}}^{\text{V}}$}  \\ \hline
			&\multicolumn{2}{c|}{C\&W} &\multicolumn{2}{c|}{PGD} &\multicolumn{2}{c|}{I-FGSM} &\multicolumn{2}{c||}{MI-FGSM} &\multicolumn{4}{c|}{$\text{DI}^2\text{-FGSM}$}\\ \hline
			$c$  &\tiny{{ASR}$_{\text{TN}}$}    &\tiny{PSNR}   &\tiny{{ASR}$_{\text{TN}}$}    &\tiny{PSNR}   &\tiny{{ASR}$_{\text{TN}}$}    &\tiny{PSNR}     &\tiny{{ASR}$_{\text{TN}}$}    &\tiny{PSNR}  &\scriptsize{iter} &\tiny{{ASR}$_{\text{SN}}$} &\tiny{{ASR}$_{\text{TN}}$}    &\tiny{PSNR}  \\ \hline
			0      &0.2    &72.0    &0.2  &74.5    &4.8   &59.7    &4.8   &59.7    &1    &91.0    &41.8    &48.2\\ \hline
			50     &60.0   &52.2    &61.2 &50.3    &65.0  &48.9    &65.4  &48.8    &2    &99.0    &66.4    &44.6\\ \hline
			80     &82.0   &47.8    &84.4 &45.8    &88.0  &45.1    &88.0  &44.8    &3    &99.6    &79.4    &42.4\\ \hline
			100    &95.0   &45.2    &96.6 &43.5    &97.4  &42.9    &97.4  &42.7    &5    &100     &92.4    &40.2\\ \hline
\multicolumn{13}{|c|}{SN = $\phi_{\text{VGGnet}}^{\text{R}}$,  TN = $\phi_{\text{VGGnet}}^{\text{V}}$}  \\ \hline
			&\multicolumn{2}{c|}{C\&W} &\multicolumn{2}{c|}{PGD} &\multicolumn{2}{c|}{I-FGSM} &\multicolumn{2}{c||}{MI-FGSM} &\multicolumn{4}{c|}{$\text{DI}^2\text{-FGSM}$}\\ \hline
			$c$  &\tiny{{ASR}$_{\text{TN}}$}    &\tiny{PSNR}   &\tiny{{ASR}$_{\text{TN}}$}    &\tiny{PSNR}   &\tiny{{ASR}$_{\text{TN}}$}    &\tiny{PSNR}     &\tiny{{ASR}$_{\text{TN}}$}    &\tiny{PSNR}  &\scriptsize{iter} &\tiny{{ASR}$_{\text{SN}}$} &\tiny{{ASR}$_{\text{TN}}$}    &\tiny{PSNR}  \\ \hline
			0      &0.4   &69.1    &0.4  &67.3    &7.8   &58.1   &7.8   &58.1 &1 &87.0 &58.2 &48.3\\ \hline
			11     &31.8  &59.2    &31.6 &56.2    &74.4  &52.0   &76.4  &51.7 &2 &96.8 &80.0 &45.0\\ \hline
			11.5   &59.8  &57.3    &66.2 &54.4    &89.0  &50.2   &89.4  &50.0 &3 &98.6 &91.8 &43.2\\ \hline
			12     &91.6  &54.9    &93.0 &52.0    &96.6  &47.9   &96.6  &47.8 &5 &100  &97.8 &41.4\\ \hline
	\end{tabular}}
	\label{tab:crosstraining1}
\vspace{-0.2cm}
\end{table}

The results for cross-training mismatch for the combinations of (SN,TN) = $(\phi_{\text{BSnet}}^{\text{R}}, \phi_{\text{BSnet}}^{\text{V}})$ and (SN,TN) = $(\phi_{\text{VGGnet}}^{\text{R}}, \phi_{\text{VGGnet}}^{\text{V}})$ are reported in Tables \ref{tab:crosstraining1} and \ref{tab:crosstraining2} for the median filtering and image resizing detection tasks, respectively.
These results are in line with the previous ones and show that increasing the confidence always helps to improve the transferability of the adversarial examples. Specifically, among the confidence-based attacks, the improvement of the percentage goes from 60\% to more than 90\% for the cross-training case on the VGGnet. For other architectures, the transferability of the adversarial examples are also improved by increasing the value of $c$.
We also observe that, as with the cross-network case, transferring the attacks between networks trained for image resizing detection tends to be more difficult, with respect to the case of median filtering detection.

\begin{table}
	\renewcommand\arraystretch{1.15}
	\centering
	\setlength{\tabcolsep}{6.5pt}
	\caption{Results for cross-training mismatch for the image resizing detection.}
	{\begin{tabular}
{|m{0.4cm}<{\centering}|m{0.45cm}<{\centering}m{0.45cm}<{\centering}|m{0.45cm}<{\centering}m{0.45cm}<{\centering}|m{0.45cm}<{\centering}m{0.45cm}<{\centering}|m{0.45cm}<{\centering}m{0.45cm}<{\centering}||m{0.35cm}<{\centering}|m{0.44cm}<{\centering}m{0.44cm}<{\centering}m{0.44cm}<{\centering}|}
			\hline
\multicolumn{13}{|c|}{SN = $\phi_{\text{BSnet}}^{\text{R}}$,  TN = $\phi_{\text{BSnet}}^{\text{V}}$}  \\ \hline
			&\multicolumn{2}{c|}{C\&W} &\multicolumn{2}{c|}{PGD} &\multicolumn{2}{c|}{I-FGSM} &\multicolumn{2}{c||}{MI-FGSM} &\multicolumn{4}{c|}{$\text{DI}^2\text{-FGSM}$}\\ \hline
			$c$  &\tiny{{ASR}$_{\text{TN}}$}    &\tiny{PSNR}   &\tiny{{ASR}$_{\text{TN}}$}    &\tiny{PSNR}   &\tiny{{ASR}$_{\text{TN}}$}    &\tiny{PSNR}     &\tiny{{ASR}$_{\text{TN}}$}    &\tiny{PSNR}  &\scriptsize{iter} &\tiny{{ASR}$_{\text{SN}}$} &\tiny{{ASR}$_{\text{TN}}$}    &\tiny{PSNR}  \\ \hline
			0        &9.8    &68.3    &9.8  &66.9   &12.8  &58.7   &12.8  &58.6  &1  &97.2  &49.2  &48.2\\ \hline
			30       &23.8   &52.5    &38.6 &49.6   &53.0  &48.0   &48.6  &47.8  &2  &99.4  &72.6  &43.8\\ \hline
			40       &32.8   &48.9    &54.2 &45.7   &64.0  &44.7   &60.2  &44.6  &3  &99.8  &80.0  &41.2\\ \hline
			50       &39.2   &45.9    &59.8 &42.3   &67.2  &41.7   &64.2  &41.7  &5  &100   &82.0  &39.1\\ \hline
\multicolumn{13}{|c|}{SN = $\phi_{\text{VGGnet}}^{\text{R}}$,  TN = $\phi_{\text{VGGnet}}^{\text{V}}$}  \\ \hline
&\multicolumn{2}{c|}{C\&W} &\multicolumn{2}{c|}{PGD} &\multicolumn{2}{c|}{I-FGSM} &\multicolumn{2}{c||}{MI-FGSM} &\multicolumn{4}{c|}{$\text{DI}^2\text{-FGSM}$}\\ \hline
			$c$  &\tiny{{ASR}$_{\text{TN}}$}    &\tiny{PSNR}   &\tiny{{ASR}$_{\text{TN}}$}    &\tiny{PSNR}   &\tiny{{ASR}$_{\text{TN}}$}    &\tiny{PSNR}     &\tiny{{ASR}$_{\text{TN}}$}    &\tiny{PSNR}  &\scriptsize{iter} &\tiny{{ASR}$_{\text{SN}}$} &\tiny{{ASR}$_{\text{TN}}$}    &\tiny{PSNR}  \\ \hline
			0        &5.6    &71.5    &5.8  &75.0    &7.6   &59.3    &7.6   &59.2  &1  &32.8  &19.2  &48.2\\ \hline
			13.5     &30.0   &50.2    &26.8 &46.8    &42.6  &45.1    &39.8  &45.1  &2  &55.2  &50.6  &45.3\\ \hline
			14       &43.6   &47.9    &42.4 &44.3    &61.0  &43.0    &59.8  &43.1  &3  &73.6  &68.0  &43.5\\ \hline
			14.5     &65.2   &45.7    &66.8 &42.0    &82.8  &41.1    &83.0  &41.2  &5  &88.6  &82.6  &41.8\\ \hline
	\end{tabular}}
	\label{tab:crosstraining2}
\vspace{-0.3cm}
\end{table}

Table \ref{tab:crossmodel-training3} reports the results for the case of AWGN detection for the cross-network setting
(SN,TN) = $(\phi_{\text{VGGnet}}^{\text{R}}, \phi_{\text{BSnet}}^{\text{R}})$ and the cross-training setting
(SN,TN) = $(\phi_{\text{BSnet}}^{\text{R}}, \phi_{\text{BSnet}}^{\text{V}})$.
Notice that only two cases are reported, as the ASR$_{\text{TN}}$ is always improved by using a large confidence value.
As one can see from the table, by using large confidence values, the ASR$_{\text{TN}}$ is improved from [0-11]\% to [80-92]\% for the cross-network case, and from [0-1]\% to [88-95]\% for the case of cross-training mismatch, thus demonstrating the effectiveness of increasing the confidence to improve the transferability. Moreover, the PSNR remains pretty high in all cases.
Eventually, Table \ref{tab:crossall} shows the results for the case of cross-network-and-training (strong mismatch) for the median filtering and image resizing detection tasks when (SN,TN) = $(\phi_{\text{BSnet}}^{\text{V}}, \phi_{\text{BC+net}}^{\text{R}})$. For simplicity, only the results corresponding to $c=0$ and the largest $c$ are reported.
For the case of median filtering detection, a significant gain ($\simeq 80\%$) in the transferability is achieved by raising $c$ while maintaining good image quality ($>$ 40 dB).
For the case of image resizing, the average gain is around 30\%.

\begin{table}
	%\resizebox{\textwidth}{!}
	\renewcommand\arraystretch{1.15}
	\centering
	\setlength{\tabcolsep}{6.5pt}
	\caption{Results for the AWGN detection task, for cross-network (top)  and cross-training (bottom) mismatch.}
	{\begin{tabular} {|m{0.4cm}<{\centering}|m{0.45cm}<{\centering}m{0.45cm}<{\centering}|m{0.45cm}<{\centering}m{0.45cm}<{\centering}|m{0.45cm}<{\centering}m{0.45cm}<{\centering}|m{0.45cm}<{\centering}m{0.45cm}<{\centering}||m{0.35cm}<{\centering}|m{0.44cm}<{\centering}m{0.44cm}<{\centering}m{0.44cm}<{\centering}|}
			\hline
\multicolumn{13}{|c|}{SN = $\phi_{\text{VGGnet}}^{\text{R}}$,  TN = $\phi_{\text{BSnet}}^{\text{R}}$}  \\ \hline
			&\multicolumn{2}{c|}{C\&W} &\multicolumn{2}{c|}{PGD} &\multicolumn{2}{c|}{I-FGSM} &\multicolumn{2}{c||}{MI-FGSM} &\multicolumn{4}{c|}{$\text{DI}^2\text{-FGSM}$}\\ \hline
			$c$  &\tiny{{ASR}$_{\text{TN}}$}    &\tiny{PSNR}   &\tiny{{ASR}$_{\text{TN}}$}    &\tiny{PSNR}   &\tiny{{ASR}$_{\text{TN}}$}    &\tiny{PSNR}     &\tiny{{ASR}$_{\text{TN}}$}    &\tiny{PSNR}  &\scriptsize{iter} &\tiny{{ASR}$_{\text{SN}}$} &\tiny{{ASR}$_{\text{TN}}$}    &\tiny{PSNR}  \\ \hline
			0       &1.2   &64.2    &5.8  &60.6    &10.6  &56.2    &10.8  &56.0  &1   &36.0   &2.2   &54.0 \\ \hline
			10      &19.2  &57.8    &20.6 &54.8    &40.4  &52.1    &41.4  &51.9  &3   &64.4   &30.6  &48.6\\ \hline
			15      &52.0  &53.8    &50.8 &51.4    &79.4  &49.4    &77.8  &49.2  &5   &82.2   &48.0  &46.8\\ \hline
			20      &79.9  &49.5    &82.8 &46.8    &91.0  &45.4    &92.2  &45.4  &10  &93.2   &63.6  &42.7\\ \hline
\multicolumn{13}{|c|}{SN = $\phi_{\text{BSnet}}^{\text{R}}$,  TN = $\phi_{\text{BSnet}}^{\text{V}}$}  \\ \hline
&\multicolumn{2}{c|}{C\&W} &\multicolumn{2}{c|}{PGD} &\multicolumn{2}{c|}{I-FGSM} &\multicolumn{2}{c||}{MI-FGSM} &\multicolumn{4}{c|}{$\text{DI}^2\text{-FGSM}$}\\ \hline
			$c$  &\tiny{{ASR}$_{\text{TN}}$}    &\tiny{PSNR}   &\tiny{{ASR}$_{\text{TN}}$}    &\tiny{PSNR}   &\tiny{{ASR}$_{\text{TN}}$}    &\tiny{PSNR}     &\tiny{{ASR}$_{\text{TN}}$}    &\tiny{PSNR}  &\scriptsize{iter} &\tiny{{ASR}$_{\text{SN}}$} &\tiny{{ASR}$_{\text{TN}}$}    &\tiny{PSNR}  \\ \hline
			0        &0.2    &65.0    &0.2  &62.4   &0.6   &57.7   &0.6   &57.6  &1   &51.2  &0.6   &54.1\\ \hline
			20       &12.2   &54.4    &11.0 &52.2   &13.8  &49.7   &15.0  &49.6  &10  &80.0  &4.8   &43.9\\ \hline
			30       &78.0   &49.4    &77.2 &47.3   &54.4  &44.8   &72.8  &45.1  &20  &91.6  &10.2  &40.0\\ \hline
			40       &95.4   &45.5    &94.0 &43.0   &88.2  &41.0   &93.0  &41.3  &30  &98.4  &21.4  &37.4\\ \hline
	\end{tabular}}
	\label{tab:crossmodel-training3}
\vspace{-0.3cm}
\end{table}

With regard to the comparison with $\text{DI}^2\text{-FGSM}$, we observe that similar ASR$_{\text{TN}}$ are achieved in most cases with lower PSNR values. The gain in PSNR is particularly evident in the case of AWGN detection.
Given that $\text{DI}^2\text{-FGSM}$ is based on input processing diversity, it is not surprising that its effectiveness depends heavily on the specific forensic task.
Notably, $\text{DI}^2\text{-FGSM}$ and the confidence-based method proposed in this paper are different and somewhat complementary approaches, that could be also combined together to further increase attack transferability.
Finally, we verified that, similarly to what happens passing from I-FGSM to MI-FGSM with PSNR limitation, integrating the momentum method in $\text{DI}^2\text{-FGSM}$ does not improve the transferability significantly.

\begin{table}
	\renewcommand\arraystretch{1.15}
	\centering
	\setlength{\tabcolsep}{6.5pt}
	\caption{Results for cross-network-and-training mismatch for the median filtering (top) and image resizing (bottom) detection tasks.}
	{\begin{tabular}
{|m{0.4cm}<{\centering}|m{0.45cm}<{\centering}m{0.45cm}<{\centering}|m{0.45cm}<{\centering}m{0.45cm}<{\centering}|m{0.45cm}<{\centering}m{0.45cm}<{\centering}|m{0.45cm}<{\centering}m{0.45cm}<{\centering}||m{0.35cm}<{\centering}|m{0.44cm}<{\centering}m{0.44cm}<{\centering}m{0.44cm}<{\centering}|}
			\hline
\multicolumn{13}{|c|}{SN = $\phi_{\text{BSnet}}^{\text{V}}$,  TN = $\phi_{\text{BC+net}}^{\text{R}}$}  \\ \hline
			&\multicolumn{2}{c|}{C\&W} &\multicolumn{2}{c|}{PGD} &\multicolumn{2}{c|}{I-FGSM} &\multicolumn{2}{c||}{MI-FGSM} &\multicolumn{4}{c|}{$\text{DI}^2\text{-FGSM}$}\\ \hline
			c  &\tiny{{ASR}$_{\text{TN}}$}    &\tiny{PSNR}   &\tiny{{ASR}$_{\text{TN}}$}    &\tiny{PSNR}   &\tiny{{ASR}$_{\text{TN}}$}    &\tiny{PSNR}     &\tiny{{ASR}$_{\text{TN}}$}    &\tiny{PSNR}  &\scriptsize{iter} &\tiny{{ASR}$_{\text{SN}}$} &\tiny{{ASR}$_{\text{TN}}$}    &\tiny{PSNR}  \\ \hline
			0     &0.0   &70.5   &0.0  &71.9   &2.6  &60.0  &2.6    &60.0   &1  &95.4  &35.6  &48.2\\ \hline
			100   &78.0  &45.1   &83.0 &43.1   &84.6 &42.4  &85.6   &42.4   &5  &100   &82.0  &40.5\\
			\hline\hline
            0     &0.8   &73.3  &0.8  &74.7  &2.0   &59.8  &2.0   &59.7     &1  &80.6  &0.0   &48.2 \\ \hline
			400   &17.0  &33.6  &40.0 &31.2  &37.4  &31.2  &33.4  &31.1     &20 &100   &26.0  &31.0\\ \hline
	\end{tabular}}
	\label{tab:crossall}
\vspace{-0.3cm}
\end{table}

\section{Discussion and Conclusions}\label{sec:conclusion}

Following some works indicating a certain lack of transferability of CNN adversarial examples in image forensic applications, we introduced a general strategy to control the strength of the attacks based on the margin between the logit value of the target class and those of the other classes. Based on our experiments, we can conclude that
by increasing the confidence margin, the attacks can be transferred  in most of the cases (ASR $>$ 80\%), regardless of the specific attacking algorithm, while the PSNR of the attacked images remains good ($>$ 30 dB).
In some cases, a slightly larger distortion is necessary to get high transfer rates,
the achievable transferability (given a minimum PSNR for the attack) depending on the detection task and the model targeted by the attack.
Future research will focus on the use of the increased-confidence attack to evaluate the security of existing defences against adversarial examples, e.g. those based on randomization strategies \cite{barni2020random,taran2019defending}, and to develop new more powerful defence mechanisms.


\section*{Acknowledgment}
This work was supported in part by the National Key Research and Development of China (2016YFB0800404), National NSF of China (U1936212, 61672090), and in part by the PREMIER project, funded by the Italian Ministry of Education, University, and Research (MIUR) within the PRIN 2017  2017Z595XS-001 program.

\bibliographystyle{splncs04}
\bibliography{ref}

\end{document}